\documentclass{article}

\usepackage{microtype}
\usepackage{subfigure}
\usepackage{booktabs} 
\usepackage{caption} 
\usepackage{subcaption}
\usepackage[pdftex]{graphicx}

\usepackage{hyperref}

\usepackage[accepted]{icml2024}

\usepackage{amsmath}
\usepackage{amssymb}
\usepackage{mathtools}
\usepackage{amsthm}
\usepackage{url}            
\usepackage{booktabs}       
\usepackage{amsfonts}       
\usepackage{nicefrac}       
\usepackage{microtype}      
\usepackage{xcolor}         
\usepackage{adjustbox}
\usepackage{multirow}
\usepackage{bbm}
\usepackage[euler]{textgreek}
\usepackage{pifont}
\usepackage{svg}
\usepackage{wrapfig}
\usepackage{pgfsys}

\usepackage[tableposition=top]{caption}
\usepackage[capitalize,noabbrev]{cleveref}

\theoremstyle{plain}

\theoremstyle{definition}

\theoremstyle{remark}

\usepackage[textsize=tiny]{todonotes}

\icmltitlerunning{EquiAV: Leveraging Equivariance for  Audio-Visual Contrastive Learning}

\begin{document}

\twocolumn[

\icmltitle{EquiAV: Leveraging Equivariance for  Audio-Visual Contrastive Learning}



\icmlsetsymbol{equal}{*}

\begin{icmlauthorlist}
\icmlauthor{Jongsuk Kim}{equal,yyy}
\icmlauthor{Hyeongkeun Lee}{equal,yyy}
\icmlauthor{Kyeongha Rho}{equal,yyy}
\icmlauthor{Junmo Kim}{yyy}
\icmlauthor{Joon Son Chung}{yyy}
\end{icmlauthorlist}

\icmlaffiliation{yyy}{Korea Advanced Institute of Science and Technology (KAIST), Daejeon, Republic of Korea}

\icmlcorrespondingauthor{Jongsuk Kim}{jskpop@kaist.ac.kr}

\icmlkeywords{Machine Learning, ICML}

\vskip 0.3in 
]



\printAffiliationsAndNotice{\icmlEqualContribution} 

\begin{abstract}
Recent advancements in self-supervised audio-visual representation learning have demonstrated its potential to capture rich and comprehensive representations. However, despite the advantages of data augmentation verified in many learning methods, audio-visual learning has struggled to fully harness these benefits, as augmentations can easily disrupt the correspondence between input pairs. To address this limitation, we introduce EquiAV, a novel framework that leverages equivariance for audio-visual contrastive learning. Our approach begins with extending equivariance to audio-visual learning, facilitated by a shared attention-based transformation predictor. It enables the aggregation of features from diverse augmentations into a representative embedding, providing robust supervision. Notably, this is achieved with minimal computational overhead. Extensive ablation studies and qualitative results verify the effectiveness of our method. EquiAV outperforms previous works across various audio-visual benchmarks. The code is available on \href{https://github.com/JongSuk1/EquiAV}{https://github.com/JongSuk1/EquiAV}
\end{abstract}

\section{Introduction}

Audio and visual modalities play a pivotal role in how humans perceive their surroundings. 
Despite the differences in their characteristics, there exists an inherent correspondence between the two modalities. 
Learning such audio-visual correspondence from large-scale unlabeled video data in a self-supervised manner has recently become a major interest in the deep-learning research community. 
Among various approaches for audio-visual self-supervised learning, \textit{Audio-Visual Contrastive Learning} has been popular due to its simplicity and effectiveness~\cite{ ma2021active,ma2021contrastive,morgado2021robust,morgado2021audio,Patrick2021ICCV,recasens2021broaden,wang2021multimodal,sarkar2023self}. 
\begin{figure}[t]
    \centering
    \includegraphics[width=0.47\textwidth]{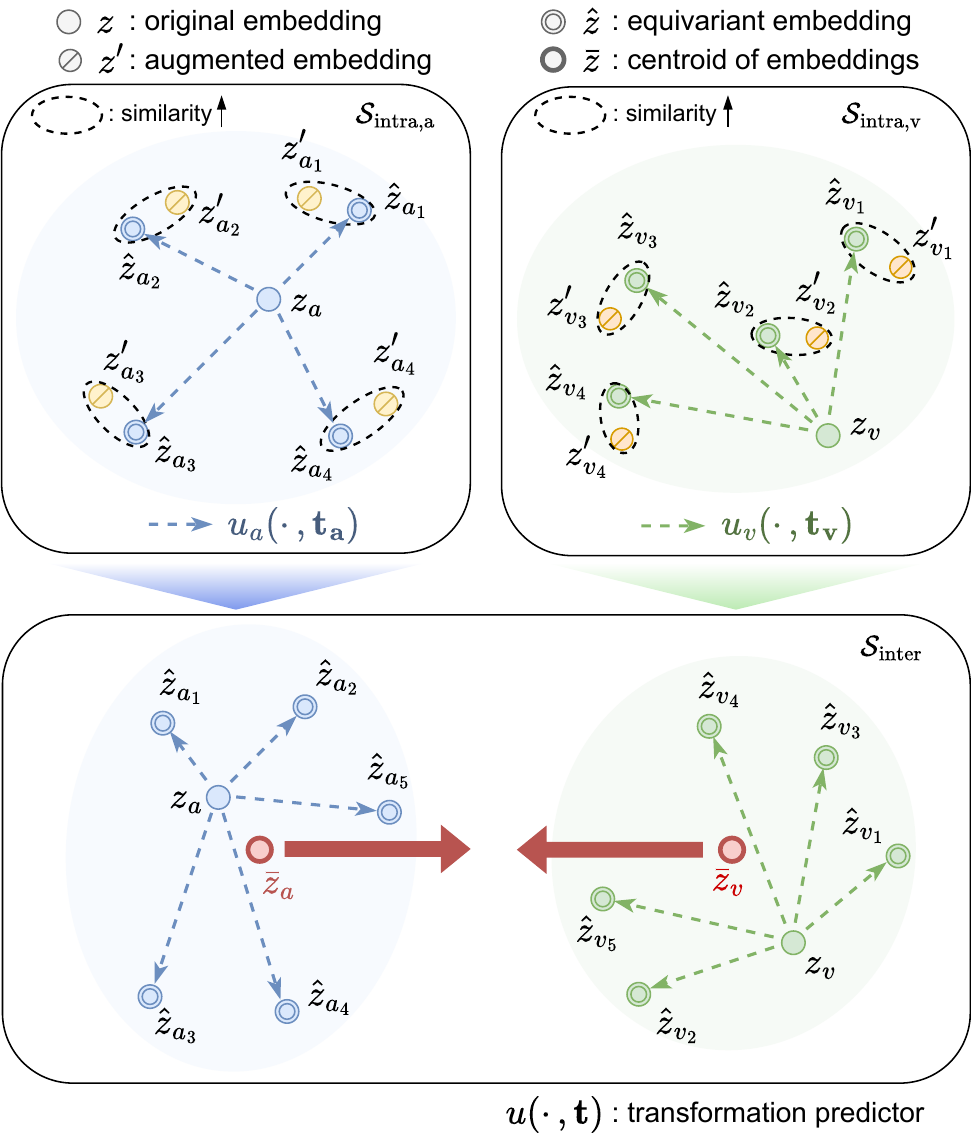}
     \label{fig:2_1_1}
    \caption{Conceptual illustration of EquiAV. Within the intra-modal latent space, the model learns augmentation-related information by leveraging equivariance. Extending equivariance to the inter-modal space provides robust cross-modal supervision.}
    \label{fig:2_1}
    \vspace{-1.5em}
\end{figure}

A key challenge in audio-visual contrastive learning is to enhance representational capability and diversity while maintaining the correspondence between two different modalities.
The most straightforward and effective way to enrich the representation is to utilize data augmentation.
However, data augmentation has not been extensively applied in the field of audio-visual contrastive learning. 
This is because correspondence between audio and visual modalities can be easily distorted due to data augmentations.
For this reason, previous works utilize a very limited range of data augmentation and adopt alternatives to learn richer representations, such as incorporating masked data modeling~\cite{gong2023contrastive,huang2023mavil} or utilizing temporal supervision~\cite{sarkar2023self,jenni2023audio}.

Meanwhile, there have been several recent studies~\cite{dangovski2022equivariant,devillers2023equimod,sie2023icml} on learning single-modal representations by employing equivariance, which can be complementary to learning representations that are invariant to data augmentations.
Equivariant latent space learns to capture augmentation-related information, thereby enhancing the representational capability.
In Particular, EquiMod~\cite{devillers2023equimod} proposes the transformation predictor to estimate the displacement in the latent space that corresponds to the transformation applied to the input space.

In this paper, we analyze the impact of applying equivariance to the learning audio-visual correspondence and joint representations and then discuss the optimal way to incorporate it into the self-supervised audio-visual contrastive learning framework.
We demonstrate that applying the equivariance in the intra-modal latent space is helpful for learning rich modality-unique information, thereby enhancing representational capability.
However, it still cannot directly resolve the negative effects of distorted audio-visual correspondence caused by augmentations.

To address this issue, we propose EquiAV, a novel framework that efficiently leverages equivariance for audio-visual contrastive learning. 
Firstly, we discover that equivariant representation learned in the intra-modal latent space can be transferred to the inter-modal latent space via a shared transformation predictor.
Taking advantage of this characteristic, we compute the centroid of a set of equivariant embeddings generated by the transformation predictor for each modality and use it for inter-modal contrastive learning.
We verify that supervision from centroids improves representational capability while reducing the undesirable effect of augmentations.
This approach requires minimal additional computational cost since the equivariant embeddings are produced from the single original inputs instead of encoding all augmented inputs.
Besides, we devise the attention-based transformation predictor that accurately encodes the parameterized augmentation vector to the latent space.
We demonstrate the effectiveness of each component of our framework through extensive ablation studies.
EquiAV outperforms the existing state-of-the-art audio-visual self-supervised pre-training methods in diverse downstream tasks, including audio-visual event classification and zero-shot audio-visual retrieval tasks.

The contribution of our paper is summarized as follows:
\begin{itemize}
\item We propose EquiAV, a novel framework that incorporates equivariance to audio-visual contrastive learning.
\item We effectively transfer equivariant representations from the intra-modal latent space to the inter-modal latent space with a shared transformation predictor. Using the centroid of equivariant embeddings enables the model to learn rich joint representations while avoiding the adverse effect of augmentations.
\item We design the attention-based transformation predictor to encode the parameterized augmentation vector into the latent space.
\item EquiAV outperforms existing audio-visual self-supervised learning methods in audio-visual event classification and zero-shot audio-visual retrieval tasks. 
\end{itemize}

\section{Related Works} \label{sec:related}
\paragraph{Audio-Visual Representation Learning.}

Audio-visual contrastive learning has been one of the most popular approaches for learning the natural correspondence between audio and visual modalities, due to its simple intuition and powerful performance on the downstream tasks~\cite{ma2021active,ma2021contrastive,morgado2021robust,morgado2021audio,Patrick2021ICCV, recasens2021broaden,wang2021multimodal,owens2018audio,sarkar2023self}.
A common approach to audio-visual contrastive learning is to learn the context of the synchronous relationship between audio and visual inputs~\cite{korbar2018coop, alwassel2020self,morgado2021audio,sarkar2023self}. 
On the other hand, several works~\cite{Georgescu2023ICCV, haliassos2022jointly} adopt masked modeling techniques that are designed to reconstruct the masked raw inputs or predict the masked context features.
Recently, CAV-MAE~\cite{gong2023contrastive} and MAViL~\cite{huang2023mavil} have combined contrastive learning and masked data modeling techniques to learn complementary representations.

\paragraph{Self-supervised Equivariant Representation Learning.}
Self-supervised learning~\cite{chen2020simple, he2020momentum, oord2018representation,he2022mae, tong2022videomae, huang2022masked,caron2021emerging, chen2021exploring, grill2020bootstrap} with large-scale datasets has demonstrated promising performance across various domains.
Many of these methods employ a joint-embedding framework, where the objective is to maximize the similarity between the embeddings of two augmented views derived from a single input.
They focus on learning representations that are invariant to such augmentations.
Recently, several studies~\cite{lee2021improving,dangovski2022equivariant,devillers2023equimod,sie2023icml} have illustrated that models can achieve better representation learning by incorporating the principle of equivariance.
Equivariance ensures that the semantic contents of representations adapt in response to the input data augmentation, thereby effectively capturing augmentation-related information within the representations.
One approach to learning equivariance is to use the auxiliary task of predicting the specific augmentations applied to the input data~\cite{dangovski2022equivariant,lee2021improving}.
Another strategy is to model the mapping between the transformation in the latent space and the augmentations in the input space~\cite{devillers2023equimod,sie2023icml}. 

\begin{figure*}[!t]
     \centering
     \includegraphics[width=0.99\textwidth]{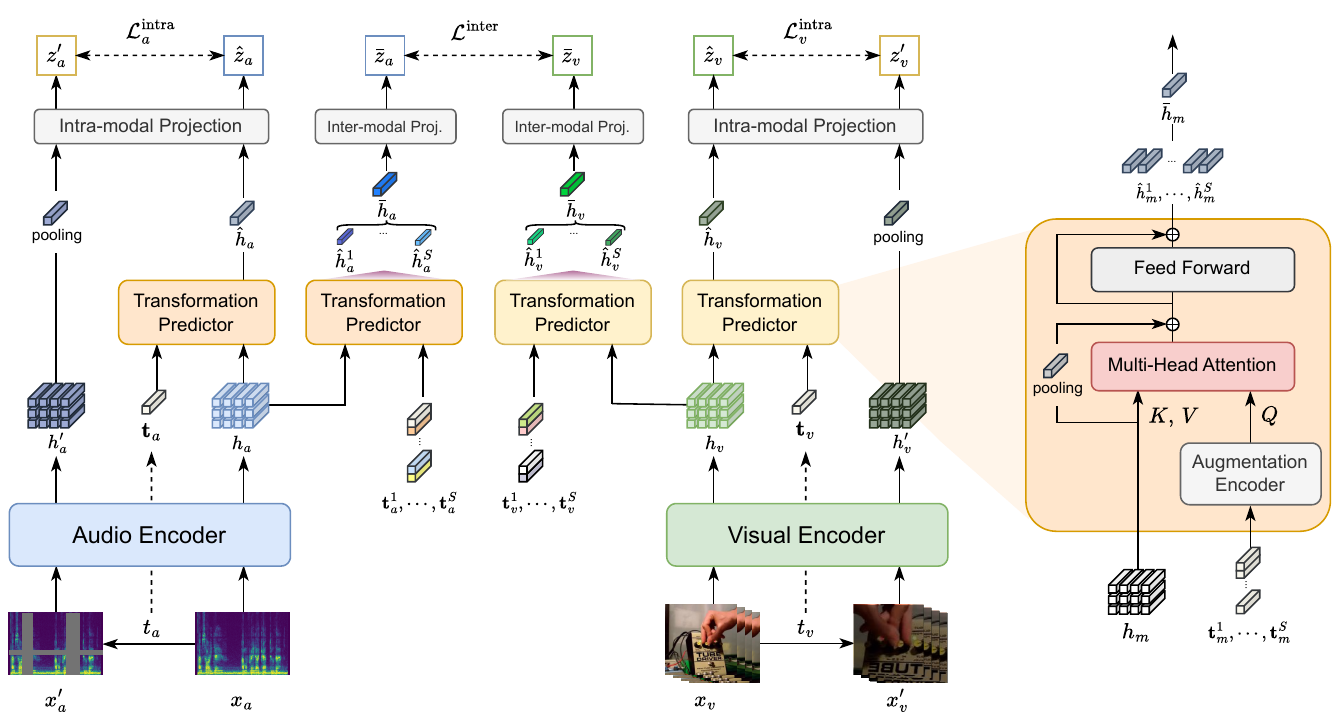}
     \caption{Overview of the proposed EquiAV framework. Given an audio-visual input pair and its augmented version, the audio encoder and the visual encoder encode them into representations. The transformation predictor takes the original representation $h_m$ and the parameterized augmentation vector $t_m$ as inputs and outputs the equivariant representation $\hat{h}_m$. In the intra-modal latent space, the model is learned to maximize the similarity of the equivariant embedding and the augmented embedding. In the inter-modal latent space, we sample multiple augmentation vectors ${\{t_{m_i}\}_{i\in\{1,...,S\}}}$ and generate the corresponding equivariant representations ${\{\hat{h}_{m_i}\}_{i\in\{1,...,S\}}}$. Then, the centroid $\bar{h}_m$ is computed and used for inter-modal contrastive learning.}
     \label{fig:model}
\end{figure*}
\section{Methods}
\label{sec:methods}

\subsection{Preliminary}
\label{sec:preliminary}

One effective approach for providing self-supervision is to utilize the concept of \textit{invariance}. 
It targets to align the representations of augmented inputs in the feature space. 
Given the encoder $f$ and the augmentation distribution $\mathcal{T}$, the objective function can be described as 
\begin{align}\label{eq:inv}
    \forall t_1,t_2 \in \mathcal{T} \quad \min_{f} \mathcal{L}(f(t_1(x), f(t_2(x))),
\end{align}
where the function $\mathcal{L}(\cdot , \cdot)$ measures the dissimilarity between two inputs.
On the other hand, some recent works~\cite{dangovski2022equivariant,devillers2023equimod, sie2023icml} have suggested that considering the discrepancy of input pairs in the feature space can lead to better representation learning. This concept, known as \textit{equivariance}, aims to capture the variations between the original input and its augmented version in a way that preserves the underlying structure. 
To implement equivariance, a transformation predictor $u$ along with augmentation $t$ is employed to match the intra-modal pair in the feature space. The objective function of equivariant self-supervised learning can be represented as follows:
\begin{equation}\label{eq:equi}
    \forall t \in \mathcal{T}, \quad \min_{u, f} \mathcal{L}(u(f(x),t), f(t(x))).
\end{equation}
Through this approach, the model can learn to encode not only the invariant features but also the specific transformations applied to input data.

\subsection{EquiAV}
\label{sec:equiav}

\paragraph{Data Augmentations and Their Parameterization.}
In our framework, typical visual-based augmentations are employed for visual inputs.
However, unlike the abundant augmentations available for an image or video, the options for the audio modality are relatively limited. 
To address this issue, we convert the audio signal into a spectrogram, followed by the application of visual-based augmentations in the same manner.
The augmentation information is encoded into real vectors, denoted as $\mathbf{t}_a$ and $\mathbf{t}_v$. These augmentation vectors parameterize how much each augmentation is applied to the input data.
A detailed explanation of data augmentation parameterization is provided in Appendix~\ref{sup:aug_param}.

\paragraph{Encoding Audio-Visual Representations.}
Given an audio-visual input pair $(x_a, x_v)$ and its augmented version $(x'_a,x'_v)$, the audio encoder $f_a$ and the visual encoder $f_v$ encode them into representations. 
The encoding process can be represented as follows:
\begin{align}
    & h_m=f_m(x_m),\quad h'_{m}=f_m(x'_m)=f_m(t_m(x_m)) 
\end{align}
where $m\in\{a,v\}$ and $t_m$ refers to the data augmentation applied to the input data. 
Note that $a$ and $v$ represent audio and visual modalities, respectively. 
We employ Vision Transformer~\cite{dosovitskiy2021an} for both encoders.

\paragraph{Equivariant Transformation.}
To predict the displacement in the latent space caused by the transformation in the input data space, a transformation predictor $u(\cdot)$ is required.
In this paper, we design the architecture of $u(\cdot)$ based on the attention mechanism.
First of all, we define an augmentation vector $\mathbf{t} \in \mathbb{R}^{S \times d_t}$ where $S$ and $d_t$ denote the number and the dimension of data augmentations $t$, respectively.
The augmentation encoder $f_{t}$ encodes the augmentation vector into the feature so that it can serve as the query of the attention layer.
The representation $h$ of the original input serves as the key and value of the attention layer to obtain the equivariant representation $\hat{h}$ as follows:
\begin{align}
\begin{split}
    \hat{h} &= u(h,\mathbf{t}) \\
    &= \operatorname{FFN}(\operatorname{MHA}(f_{t}(\mathbf{t}),h,h) + \operatorname{MeanPool}(h)),
\end{split}
\end{align}
\begin{align}
\begin{split}
\operatorname{MHA}(\mathbf{q}, \mathbf{k}, \mathbf{v}) =\operatorname{Concat}\left(\mathbf{o}_1, \ldots, \mathbf{o}_H\right) w^O, \\
    \text{where} \enskip \mathbf{o}_{j} =\operatorname{Softmax}\left(\frac{\mathbf{q} w_j^Q\left(\mathbf{k} w_j^K\right)^{\top}}{\sqrt{d_H}}\right) \mathbf{v} w_j^V,
\end{split}
\end{align}
where $H$ is the number of heads, $d_H$ is dimension of each attention head, $w_h^Q, w_h^K, w_h^V \in \mathbb{R}^{d \times d_H}$, $ w^O \in \mathbb{R}^{H d_H \times d}$, and $j \in \{1,...,H\}$.
The Multi-Head Attention (MHA) layer calculates the score through query and key to identify the relevance of the augmentation feature in a patch-wise manner. 
These values are incorporated into the value to determine the displacement within the feature space.
Then, the output of the multi-head attention layer is added to the representation of the original input image and passed through the Feed Forward Network (FFN). 

\paragraph{Intra-modal Equivariant Learning.}
The goal of intra-modal representation learning is to learn the equivariant latent space so that augmentation-related information can be passed to inter-modal representation learning.
To this end, the equivariant representation $\hat{h}_m$ and representation $h'_m$ of augmented input are projected into the intra-modal latent space to get the equivariant embedding $\hat{z}_m$ and the augmented embedding $z'_m$ as follows:  
\begin{align}
    & \hat{z}_m = g_m^\text{intra}(\hat{h}_m) = g_m^\text{intra}(u_m(h_m,\mathbf{t_m})) \\
    & z'_m=g_m^\text{intra}(\operatorname{MeanPool}(h'_m)),
\end{align}
where $g_m^\text{intra}$ denotes the intra-modal projection head.

For intra-modal equivariant contrastive learning,
the batch consisting of $N$ equivariant embeddings $\{\hat{z}_{m}^{i}\}_{i\in \{1, \ldots, N\}}$ obtained by incorporating the augmentation information and original inputs, as well as $N$ embeddings $\{{z'}_{m}^{i}\}_{i\in \{1, \ldots, N\}}$ of the augmented inputs is used.
The pair $(\hat{z}_{m}^{i}, {z'}_{m}^{i})$ generated from the same image forms a positive pair, while the remaining $2(N-1)$ embeddings within the batch serve as negative pairs. The set of embeddings for the negative pairs can be represented as follows:
\begin{equation}
    {{Z_m^{i^{-}}}} = \left\{\bigcup_{j=1}^{N} \{\hat{z}_{m}^{j}\} \setminus \{\hat{z}_{m}^{i}\}\right\} \cup \left\{\bigcup_{j=1}^{N} \{ {z'}_{m}^{j}\} \setminus \{ {z'}_{m}^{i}\}\right\}.
\end{equation}
The NT-Xent loss~\cite{chen2020simple} is employed to compute the contrastive loss. The loss for equivariant embeddings of the augmented input can be expressed as follows:
\begin{equation}\label{loss:intra_ell}
    \resizebox{0.95\hsize}{!}{$
    \displaystyle\ell^{\text{intra}}(\hat{z}_m, z'_m) = -\frac{1}{N} \sum_{i=1}^{N} \log \frac{\exp (\text{sim}(\hat{z}_{m}^{i}, {z'}_{m}^{i})/\tau)}{\sum_{z\in {Z_m^{i^{-}}}\cup\{{z'}_{m}^{i}\}} \exp(\text{sim}(\hat{z}_{m}^{i}, z/\tau)},
    $}
\end{equation}
where $\tau > 0$ is temperature. 
Then, the intra-modal loss is described by combining the losses of adopting the original and augmented inputs as anchors respectively,
\begin{equation}\label{loss:intra}
   \mathcal{L}^{\text{intra}} = \frac{1}{2} (\ell^{\text{intra}}(\hat{z}_m, z'_m) + \ell^{\text{intra}}(z'_m, \hat{z}_m)).
\end{equation}
The process of aligning these two embeddings not only enhances the representation capability but also benefits in acquiring a representative vector for inter-modal correspondence learning.

\paragraph{Inter-modal Contrastive Learning.}
The most important factor in learning inter-modal correspondences is how to draw representative vectors from input data.
A basic approach is to use the embedding of the original data as the representative value.
It has the advantage of maximizing the correspondence information of the given data but suffers from the lack of input diversity.
On the other hand, using the embedding of augmented input strengthens the representation capability, but may disrupt the correspondence of the given pair.
Then, we can think of the expectation value of $h'$ over the set $D_h$ of representations of augmented inputs as follows:
\begin{equation}
    h^{\text{rep}}=\mathbb{E}_{h'\sim D_h}\left[ h'\right].
\end{equation}
The vector $h^{\text{rep}}$ integrates information from multiple augmented data, securing more general features while also providing robustness against undesirable effects of augmentation. 
However, it is not feasible to generate all transformations and obtain the representative vector $h^{\text{rep}}$, we aim to approximate it with the centroid of sampled equivariant representations.
\begin{align}\displaystyle
    h^{\text{rep}}_m \approx \bar{h}_m &= \frac{1}{S}\sum_{i=1}^{S} \hat{h}_{m}^{i} = \frac{1}{S}\sum_{i=1}^{S} u_m(h_{m}^{i},\mathbf{t}_{m}^{i}).
\end{align}
Note that the weight of $u_m$ is shared between the intra-modal transformation predictor and the inter-modal transformation predictor.
This allows the transfer of equivariance learned in the intra-modal space to the inter-modal space.
Furthermore, since we use equivariant representations instead of augmented representations generated by passing through the encoder, there is minimal impact on computational cost. 
Lastly, the centroid is projected into the inter-modal latent space via the projection head $g_a^\text{inter}$ and $g_v^\text{inter}$ as follows:
\begin{align}\displaystyle
    \bar{z}_m &= g_m^{\text{inter}}(\bar{h}_m).
\end{align}
In the training phase, the batch consisting of $N$ paired embeddings $\{(\bar{z}^{i}_{a}, \bar{z}^{i}_{v})\}_{i\in \{1, \ldots, N\}}$, which are extracted without applying any augmentation to preserve the correspondence between input pairs, is used.
Then, the contrastive loss for audio-visual learning can be expressed as follows:
\begin{equation}
    \ell^{\text{inter}}(\bar{z}_a, \bar{z}_v) = -\frac{1}{N} \sum_{i=1}^{N} \log \frac{\exp (\text{sim}(\bar{z}^{i}_{a}, \bar{z}^{i}_{v})/\tau)}{\sum_{j=1}^{N} \exp(\text{sim}(\bar{z}^{i}_{a}, \bar{z}^{j}_{v})/\tau)},
\end{equation}
where $\tau > 0$ is temperature. Then, the inter-modal loss is represented by combining the losses of adopting the audio and visual embeddings as anchors respectively,
\begin{equation}
\label{loss:inter}
    \mathcal{L}^{\text{inter}} = \frac{1}{2} (\ell^{\text{inter}}(\bar{z}_a, \bar{z}_v) + \ell^{\text{inter}}(\bar{z}_v,\bar{z}_a)).
\end{equation}

Finally, the loss function of EquiAV can be expressed by incorporating scaling factors, it can be formulated as follows:
\begin{equation}\label{loss:equiav}
    \mathcal{L}_{\text{EquiAV}} = \lambda^{\text{inter}}  \mathcal{L}^{\text{inter}} + \lambda^{\text{intra}}_a  \mathcal{L}^{\text{intra}}_a + \lambda^{\text{intra}}_v  \mathcal{L}^{\text{intra}}_v,
\end{equation}
where $\mathcal{L}^{\text{intra}}_a$ and $\mathcal{L}^{\text{intra}}_v$ denote intra-modal loss for audio and visual modalities, respectively.

Through these methods, EquiAV can maximize the benefits of equivariant representation learning while avoiding any detrimental impact on the learning of inter-modal pairs. 
In the next section, we will provide experimental evidence supporting our design choices along with benchmark performance. 
Algorithm~\ref{alg} summarizes our method.

\begin{algorithm}[t!]
\small
\caption{EquiAV}
\label{alg}
    \begin{algorithmic}
        \STATE {\bfseries Input:} backbone encoder $f_a$, $f_v$,
        \STATE intra-modal projection head $g_a^\text{intra}$, $g_v^\text{intra}$,
        \STATE inter-modal projection head $g_a^\text{inter}$, $g_v^\text{inter}$, 
        \STATE augmentation predictor $u_a$, $u_v$,
        \STATE audio, visual augmentation distribution $\mathcal{T}_{a}$, $\mathcal{T}_{v}$
        \STATE weight scaling factors $\lambda^{\text{inter}}$, $\lambda^{\text{inter}}_{a}$, $\lambda^{\text{inter}}_{v}$
        \STATE batch size $N$, temperature $\tau$
        \FOR{sampled mini-batch $\{(x_a^k,x_v^k)\}_{k=1}^N$} 
            \FOR{$m\in\{a,v\}$}
                \FOR{\textbf{all} $k\in \{1, \ldots , N\}$}

                    \STATE sample augmentation vectors $\mathbf{t}_m^\text{intra} \sim \mathcal{T}_{m}$, 
                    \STATE $h_m^k = f_m(x_m^k), \quad {h'}_{m}^{k} = f_m(t_m^\text{intra}(x_m^k))$ 
                    \STATE $\hat{h}_m^k= u_m(h_m^k,\mathbf{t}_{m}^\text{intra})$
                    \STATE $z_m^{2k-1} = g_m^\text{intra}(\hat{h}_m^k) $ 
                    \STATE $z_m^{2k} = g_m^\text{intra}(\operatorname{MeanPool}({h'}_{m}^{k}))$
                    
                    \STATE sample augmentation vectors $\{\mathbf{t}_{m_{i}}^\text{inter}\}_{i\in\{1,...,S\}} \sim \mathcal{T}_{m}$, 
                    \STATE $\mathbf{t}_{m}^\text{inter} = \begin{bmatrix} t_{m_1}^\top,  t_{m_2}^\top, \cdots, t_{m_S}^\top \end{bmatrix}^{\top}$
                    \STATE $\bar{h}_m^k = \operatorname{MeanPool}(u_m(h_m^k,\mathbf{t}_{m}^\text{inter}))$
                    \STATE $\bar{z}_m^k = g_m^\text{inter}(\bar{h}_m^k)$
                \ENDFOR
            
                \FOR{\textbf{all} $i \in \{1,\ldots,2N\}$ and $j \in \{1,\ldots,2N\}$}
                    \STATE $s_m^{i,j} = \exp \left({z_m^{i\top}} z_m^j / (\tau \|z_m^i\| \|z_m^j\|) \right) $

                \ENDFOR
                \STATE \textbf{define} $\ell_m^{\text{intra}}(i,j)$ \textbf{as} 
                \STATE $\ell_m^{\text{intra}}(i,j)=-\log \left( \frac{s_m^{i,j}}{\sum_{k=1}^{2N} \mathbbm{1}_{[k \neq i]} s_m^{i,k}}\right)$
                \STATE $ \mathcal{L}_m^{\text{intra}} = \frac{1}{2N}$ \scalebox{0.9}{$\sum_{k=1}^{N}\left[ \ell_{m}^{\text{intra}}(2k-1,2k) + \ell_{m}^{\text{intra}}(2k, 2k-1)\right]$}
            \ENDFOR
            \FOR{\textbf{all} $i \in \{1,\ldots,N\}$ and $j \in \{1,\ldots,N\}$}
                \STATE $s^{i,j} = \exp \left(\bar{z}^{i\top}_a \bar{z}^j_v / (\tau \|\bar{z}^i_a\| \|\bar{z}^j_v\|) \right) $
            \ENDFOR
            \STATE \textbf{define} $\ell^{\text{inter}}(i)$ \textbf{as} 
            \STATE $\ell^{\text{inter}}(i)=-\frac{1}{2}\left(\log \left( \frac{s^{i,i}}{\sum_{k=1}^{N} s^{i,k}}\right) + \log \left( \frac{s^{i,i}}{\sum_{k=1}^{N} s^{k,i}}\right)\right)$
            \STATE $\mathcal{L}^{\text{inter}} = \frac{1}{N} \sum_{i=1}^{N} \ell^{\text{inter}}(i)$
            \STATE $\mathcal{L} = \lambda^{\text{inter}}\mathcal{L}^{\text{inter}} + \lambda_{a}^{\text{intra}}\mathcal{L}_{a}^{\text{intra}} + \lambda_{v}^{\text{intra}}\mathcal{L}_{v}^{\text{intra}}$
            \STATE update networks to minimize $\mathcal{L}$
        \ENDFOR  
    \end{algorithmic}
\end{algorithm}
\section{Experiments}
\begin{table*}[ht]
\setlength{\tabcolsep}{5pt}
    \caption{Audio-visual event classification performance on AudioSet and VGGSound. \textbf{A}: Audio-only, \textbf{V}: Visual-only, \textbf{A-V}: Audio-visual. IN SL: ImageNet supervised learning, SSL: Self-supervised learning, $^*$ Non-standard train/test split. }
    \vspace{-0.5em}
    \label{exp:main}
    \centering
    \begin{adjustbox}{max width=\textwidth}
    \begin{tabular}{lccccccccccc}
        \toprule
        & \multirow{2}{*}{Pretrain} & \multicolumn{3}{c}{AudioSet-20K (mAP)} & \multicolumn{3}{c}{AudioSet-2M (mAP)} & \multicolumn{3}{c}{VGGSound (acc.)} \\ 
        \cmidrule(lr){3-5} \cmidrule(lr){6-8} \cmidrule(ll){9-11}
        Method &  & A & V & A-V & A & V & A-V & A & V & A-V \\
        \midrule
        GBlend~\citep{wang2020makes} & - & 29.1 & 22.1 & 37.8 & 32.4 & 18.8 & 41.8  & - & - & - \\
        Perceiver~\citep{jaegle2021perceiver} & - & - & - & - & 38.4 & 25.8 & 44.2 & - & - & -\\
        Attn AV~\citep{fayek2020large} & IN SL  & - & - & - & 38.4 & 25.7 & 46.2 & - & - & - \\
        MBT$^*$~\citep{nagrani2021attention} & IN21K SL & 31.3 & \textbf{27.7} & 43.9 & 41.5 & \textbf{31.3} & 49.6 & 52.3 & \textbf{51.2} & 64.1 \\
        CAV-MAE~\citep{gong2023contrastive} & SSL & 37.7 & 19.8 & 42.0 & 46.6 & 26.2 & 51.2 & 59.5 & 47.0 & 65.5 \\
        AudiovisualMAE$^*$~\citep{Georgescu2023ICCV} & SSL & - & - & - & 46.6 & 31.1 & 51.8 & 57.2 & 50.3 & 65.0 \\
        MAViL~\citep{huang2023mavil} & SSL & 41.8 & 24.8 & 44.9 & 48.7 & 30.3 & 53.3 & 60.8 & 50.9 & \textbf{67.1} \\
        EquiAV (ours) & SSL & \textbf{42.4} & 25.7 & \textbf{46.6} & \textbf{49.1} & 30.1 & \textbf{54.6} & \textbf{61.0} & 50.7 & \textbf{67.1} \\
        \bottomrule
  \end{tabular}
  \end{adjustbox}
  \vspace{-1em}
\end{table*}
The model is pre-trained on the AudioSet-2M in a self-supervised manner, without the use of labels. 
The audio encoder $f_a$ and visual encoder $f_v$ are initialized with the self-supervised pre-trained ViT-B/16 model of MAE~\cite{he2022mae}.
We employ 3-layer MLP with layer normalization for intra-modal and inter-modal projection heads.
Note that the multi-head attention layer and the feed-forward layer of the transformation predictor are similar to those used in Transformers~\cite{vaswani2017attention,chen2021crossvit}, which include layer normalization layer and residual connection.
We set $\lambda^{\text{inter}}$, $\lambda^{\text{inter}}_{a}$ and $\lambda^{\text{inter}}_{v}$ all the same as 1.
More detailed experimental settings are explained in Appendix~\ref{sup:implementation} and Appendix~\ref{sup:hyper}.

\begin{table}[t!]
    \centering
    \caption{Audio classification accuracy of finetuning on ESC-50, SPC-v1, and FSD-50K. $^{\dagger}$Results reproduced on our environment.\footnotemark} 
    \label{sup:single}
        \begin{adjustbox}{max width=0.95\columnwidth,center}
        \centering
        \begin{tabular}{lccc}
        \toprule
        Method &  ESC-50 & SPC-v1 & FSD-50K \\
        \midrule
        CAV-MAE-Scale++$^\dagger$~\citep{gong2023contrastive}  & 83.2 & 97.1 & 45.5 \\ 
        MAViL~\citep{huang2023mavil}        &    94.4  & 97.4  & - \\ 
        XKD~\citep{sarkar2023xkd}  & \textbf{96.5} &  - & 58.5 \\
        EquiAV (ours)                       & 96.0   & \textbf{97.8}  & \textbf{62.6} \\
        \bottomrule
        \end{tabular}
        \end{adjustbox}
        \vspace{-1em}
\end{table}

\begin{table}[t!]
    \centering
    \caption{Action recognition accuracy of transfer learning on UCF101, HMDB51, and Kinetics400. $^{*}$High temporal resolution with industry-level computational resources.} 
    \label{exp:single_visual}
        \begin{adjustbox}{max width=1.0\columnwidth,center}
        \centering
        \begin{tabular}{lccc}
        \toprule
        Method &  UCF101 & HMDB51 & Kinetics400 \\
        \midrule
        XDC~\citep{alwassel2020xdc}  & 85.3 &  56.0 & - \\ 
        MMV~\citep{alayrac2020self}  & 83.9 &  60.0 & - \\ 
        BraVe$^{*}$~\citep{recasens2021broaden}  & \textbf{90.0} &  63.6 & - \\ 
        AVID~\citep{morgadoavidcma}  & - & -  & 48.9 \\ 
        CrissCross~\citep{sarkar2021crisscross}  & 87.7 & 56.2  & 50.1 \\ 
        XKD~\citep{sarkar2023xkd}  & 88.4 &  62.2 & 56.5 \\ 
        EquiAV (ours) & 89.7 & \textbf{64.4} & \textbf{57.3} \\
        \bottomrule
        \end{tabular}
        \end{adjustbox}
        \vspace{-1em}
\end{table}

\subsection{Main Results}
\paragraph{Audio-Visual Event Classification.}
We evaluate the representational capability of our model through fine-tuning on multi-modal benchmark datasets, AudioSet and VGGSound, in three settings: audio-only, visual-only, and audio-visual.
During the fine-tuning process, intra-modal and inter-modal projection heads are removed and a linear classifier is added on top of encoders. 
For audio-visual fine-tuning tasks, outputs of modality-specific encoders are concatenated to form the joint representation and used as the input for the linear classifier.
As shown in Table~\ref{exp:main}, EquiAV outperforms previous methods by solely relying on contrastive learning. 
\footnotetext{Weights from \url{https://github.com/YuanGongND/cav-mae}}

\paragraph{Comparison on the Single-modal Benchmarks.}
To demonstrate the performance of our framework on the single-modal downstream tasks, we fine-tune the audio branch of EquiAV on audio classification datasets, including the ESC-50~\cite{piczak2015esc}, SPC-v1~\cite{Warden2018SpeechCA}, and FSD-50K~\cite{fonseca2022fsd50k}. 
According to the results reported in Table~\ref{sup:single}, our method shows superior performance compared to other self-supervised learning-based methods. 
We also evaluate the video branch of EquiAV on various action recognition datasets, including UCF101~\cite{soomro2012ucf101}, HMDB51~\cite{kuehne2011hmdb}, and Kinetics400~\citep{kay2017kinetics}. 
As demonstrated in Table~\ref{exp:single_visual}, our method surpasses previous SOTA audio-visual SSL methods on uni-modal visual benchmark datasets. 
The comparison is based on results from other audio-visual SSL methods that also utilize the same pre-training dataset, AudioSet-2M. 

\begin{table}[t!]
    \caption{Zero-shot audio-visual retrieval results on the MSR-VTT, AudioSet, and VGGSound. $^{\dagger}$Results reproduced on our environment. $^{*}$Models pre-trained on the much larger dataset.}
    
    \centering
    \label{result:retrieval}
    \begin{adjustbox}{width=1.0\columnwidth,center}
    \centering
    \begin{tabular}{lccccccc}
    \toprule
    & \multicolumn{3}{c}{Video-to-Audio} & \multicolumn{3}{c}{Audio-to-Video} \\
    \cmidrule(lr){2-4} \cmidrule(lr){5-7} 
    Method &  R@1 & R@5 & R@10 & R@1 & R@5 & R@10\\
    \midrule
    \small\textit{MSR-VTT} \\
    Boggust$^{*}$ &  9.3 & 20.7 & 28.8 & 7.6 & 21.1 & 28.3 \\
    Aranjelovic$^{*}$& 11.9 & 25.9 & 34.7 & 12.6 & 26.3 & 33.7 \\
    AVLnet$^{*}$ &  \textbf{17.2} & 26.6 & \textbf{46.6} & \textbf{17.8} & \textbf{35.5} & \textbf{43.6} \\
    CAV-MAE &  13.3 & \underline{29.0} & 40.5 & 7.6 & 19.8 & 30.2 \\
    EquiAV (ours) & \underline{13.8} & \textbf{31.4} & \underline{43.0} & \underline{14.4} & \underline{33.3} & \underline{43.0} \\
    \midrule
    \small\textit{AudioSet} \\
    CAV-MAE-Scale++$^\dagger$& 16.6 & 37.0 & 45.9 & 14.3 & 32.0 & 40.7 \\
    CAV-MAE  & 18.8 & 39.5 & 50.1 & 15.1 & 34.0 & 43.0  \\
    EquiAV (ours) & \textbf{30.1} & \textbf{53.3} & \textbf{62.9} & \textbf{29.6} & \textbf{53.7} & \textbf{63.1} \\
    \midrule
    \small\textit{VGGSound} \\
    CAV-MAE-Scale++$^\dagger$& 15.5 & 35.3 & 45.1 & 16.4 & 35.0 & 44.7 \\
    CAV-MAE & 14.8 & 34.2 & 44.0 & 12.8 & 30.4 & 40.3  \\
    EquiAV (ours) & \textbf{28.5} & \textbf{51.5}  & \textbf{61.7} & \textbf{28.4} & \textbf{51.8} & \textbf{62.3}\\
    \bottomrule
    \end{tabular}
    \end{adjustbox}
    \vspace{-1em}
\end{table}

\paragraph{Zero-Shot Audio-Visual Retrieval.}
Table~\ref{result:retrieval} reports the performance of zero-shot retrieval on audio-visual datasets, based on the similarity between the audio-visual embeddings.
The sample lists used in the experiments for AudioSet and VGGSound are the same as those used in CAV-MAE~\cite{gong2023contrastive}. 
Additionally, to demonstrate the robustness of our model across different datasets, we perform further zero-shot retrieval experiments on MSR-VTT. 
Our model consistently outperforms existing methods pre-trained on AudioSet and achieves comparable results to models pre-trained on HowTo100M, such as Boggust~\cite{boggust2019grounding}, Aranjelovic~\cite{arandjelovic2017look}, and AVLnet~\cite{noroozi2016unsupervised}.

\begin{table}[t!]
\setlength{\tabcolsep}{8pt}
    \caption{Zero-shot retrieval results on AudioSet and audio-visual event classification performance on AudioSet-20K with the variants of pre-training methods. \textbf{V2A}: Video-to-audio zero-shot retrieval R@1, \textbf{A2V}: Audio-to-video zero-shot retrieval R@1. \textbf{Inv.}: Invariant learning, \textbf{Equi.}: Equivariant learning, ($z_a,z_v$): Original embeddings, ($z'_a,z'_v$): Augmented embeddings, ($\hat{z}_a,\hat{z}_v$): Equivariant embeddings, ($\Bar{z}_a,\Bar{z}_v$): Centroids of equivariant representations.}
    \vspace{-0.2em}
    \label{abl:inv_equi}
    \centering
    \begin{adjustbox}{width=1.0\columnwidth,center}
    \begin{tabular}{ccccccc}
        \toprule
        & &\multicolumn{2}{c}{ZS Retrieval} & \multicolumn{3}{c}{Fine-Tuning (mAP)} \\ 
        \cmidrule(lr){3-4} \cmidrule(ll){5-7}
        Intra.   & Inter.      & V2A       & A2V       & A         & V         & A-V\\
        \midrule
        Inv.    & ($z_a,z_v$)   & 25.2   & 24.3  & 40.1     & 22.9     & 42.3 \\
        Equi.   & ($z_a,z_v$)     &   28.7    & 28.5 &   41.9 & 24.8 & 44.9 \\
        Equi.   & ($z'_a,z'_v$) & 28.1 & 27.5 & 41.5 & 24.2 & 43.9  \\
        Equi.   & ($\hat{z}_a,\hat{z}_v$) & 27.8 & 27.4 & 41.2 & 23.6 & 43.4  \\
        Equi.   & ($\Bar{z}_a,\Bar{z}_v$) & \textbf{30.1} & \textbf{29.6} & \textbf{42.4} & \textbf{25.7} & \textbf{46.6}  \\
        \bottomrule
    \end{tabular}
    \end{adjustbox}
    \vspace{-1em}
\end{table}

\subsection{Ablation Studies}
This section presents ablation studies aimed at verifying the effectiveness of our framework.
Firstly, we demonstrate the impact of employing equivariance and centroids of audio-visual embeddings produced by the transformation predictor in audio-visual contrastive learning.
Next, we explore different architectures of the transformation predictor and training strategy.
Lastly, we analyze the impact of the employed intra-modal equivariant loss function.

\paragraph{Impact of Single-modal Equivariance.}
As presented in Table~\ref{abl:inv_equi}, employing equivariance in the intra-modal latent space results in better zero-shot retrieval and fine-tuning performances compared to the invariance-based method.
This underlines the role of intra-modal equivariance in capturing the modality-unique information, which in turn facilitates the learning of audio-visual correspondence and joint representation.
When using the augmented inputs for inter-modal contrastive learning, there is a slight decrease in both zero-shot retrieval and fine-tuning performances.
The result suggests the augmentation-related information learned through intra-modal equivariance cannot entirely compensate for the distortion of audio-visual correspondence caused by the data augmentations.

On the other hand, utilizing the centroids of the equivariant embeddings in the inter-modal latent space shows both enhanced zero-shot retrieval and fine-tuning performances.
This provides two critical insights.
Firstly, the equivariant representation is successfully transferred from the intra-modal latent space to the inter-modal latent space via the shared proposed transformation predictor.
Moreover, although the individual augmented input pair may provide inaccurate supervision regarding the audio-visual correspondence, the centroid of these embeddings offers better supervision compared to the embeddings of the original input pair.
Note that we use 16 equivariant embeddings for computing the centroid in this experiment.

\begin{table}[t!]
\setlength{\tabcolsep}{10pt}
    \caption{Zero-shot retrieval results on AudioSet and audio-visual event classification performance on AudioSet-20K with the number of equivariant representations used for computing the centroid in the inter-modal latent space.}
    \vspace{-0.3em}
    \centering
    \label{abl:num_embed}
    \begin{adjustbox}{max width=\linewidth}
    \begin{tabular}{lccccc}
        \toprule
        &\multicolumn{2}{c}{ZS Retrieval} & \multicolumn{3}{c}{Fine-Tuning (mAP)} \\ 
        \cmidrule(lr){2-3} \cmidrule(ll){4-6}
        $\#$ representations     & V2A   & A2V   & A    & V     & A-V \\
        \midrule
        1  & 27.8 & 27.4 & 41.2 & 23.6 & 43.4\\
        4 & 28.3 & 29.0 & 41.4 & 24.4 & 44.9 \\
        8 & 29.4 & 29.2  & 41.9 & 24.9 & 45.5 \\
        16 & \textbf{30.1} & \textbf{29.6} &  \textbf{42.4} & \textbf{25.7} & \textbf{46.6} \\
        \bottomrule
    \end{tabular}
    \end{adjustbox}
    \vspace{-0.7em}
\end{table}

\begin{table}[t!]
\setlength{\tabcolsep}{8pt}
    \caption{Averaged cosine similarity score and Event classification performance on AudioSet-20K based on varying architectures of transformation predictor.}
    \vspace{-0.2em}
    \centering
    \label{abl:aug_pred_arch}
    \begin{adjustbox}{max width=\linewidth}
    \begin{tabular}{lccccccc}
        \toprule 
        & \multicolumn{2}{c}{Similarity} &\multicolumn{2}{c}{ZS Retrieval} & \multicolumn{3}{c}{Fine-Tuning (mAP)} \\ 
        \cmidrule(lr){2-3} \cmidrule(lr){4-5} \cmidrule(ll){6-8}
        Method & A  & V & V2A   & A2V   & A    & V     & A-V \\
        \midrule
        Linear  & 0.85& 0.87& 28.2 & 27.5 & 41.5 & 24.1 & 43.8 \\
        Hypernet.& 0.87& 0.90& 29.6 & 29.6 & 42.0 & 25.2 & 44.5\\ 
        MHA& \textbf{0.90} & \textbf{0.92} & \textbf{30.1} & \textbf{29.6} &  \textbf{42.4} & \textbf{25.7} & \textbf{46.6}  \\
        \bottomrule
    \end{tabular}
    \end{adjustbox}
    \vspace{-1em}
\end{table}

\begin{figure*}[t]
    \centering
     \subfigure[child singing]{
     \label{fig:qual_1}
     \includegraphics[height=4cm,bb=5 0 175 175]{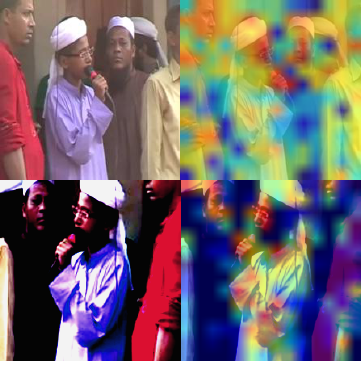}
     }
     \subfigure[playing electronic organ]
     {
     \label{fig:qual_2}
     \includegraphics[height=4cm,bb=5 0 175 175]{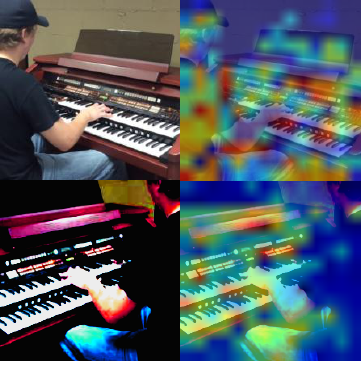}
     }
     \subfigure[playing tympani]
     {
     \label{fig:qual_3}
     \includegraphics[height=4cm,bb=5 0 175 175]{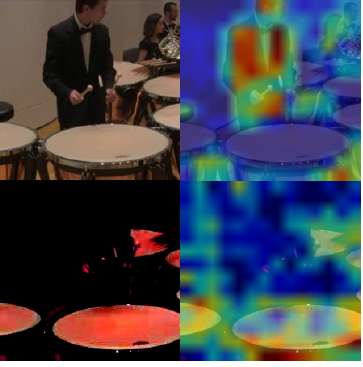}
     }
     \subfigure[children shouting]
     {
     \label{fig:qual_4}
     \includegraphics[height=4cm,bb=5 0 175 175]{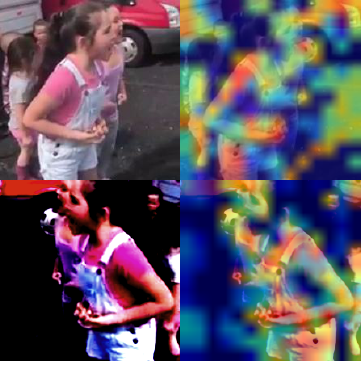}
     }
    \caption{Qualitative results of EquiAV using VGGSound. \textbf{Top Left}: Original input image, \textbf{Top Right}: Zero-shot sound source localization result for the original input image. \textbf{Bottom Left}: Augmented image whose embedding is the closest to the centroid of equivariant embeddings, \textbf{Botton Right}: Zero-shot sound source localization result for the augmented image.}
    \vspace{-1em}
    \label{fig:qual}
\end{figure*}

\paragraph{Number of Representations for Computing Centroid.}
As explained in Section~\ref{sec:equiav}, the centroid gets closer to the true mean of the representations generated from the augmented inputs by increasing the number of equivariant representations.
Table~\ref{abl:num_embed} shows that both zero-shot retrieval and fine-tuning performance consistently improve as the number of representations increases.
However, since there still exists a variance in the centroids, our method can successfully overcome the trade-off between providing accurate audio-visual correspondence and increasing the diversity of training data.
In addition, an advantage of our method is that the equivariant representations are generated from the representation of the original input using the transformation predictor, instead of passing the augmented inputs through the modality-specific encoders one by one.
This can save the significant overhead of computation and training time.
In our experiments, using 16 equivariant representations to compute the centroids increases the computational cost and training time by less than 1$\%$, as illustrated in Appendix~\ref{sup:abl}.

\paragraph{Transformation Predictor Architectures.}
We explore different architectures of the transformation predictor that models the transformation in the latent space caused by input augmentations.
Equimod~\cite{devillers2023equimod} takes a concatenation of parameterized augmentation vector and embedding as the input to a linear layer.
SIE~\cite{sie2023icml} points out that the predictor with such a simple architecture can collapse to the identity function and suggests a hypernetwork-based design that predicts the weights of the predictor.
In Table~\ref{abl:aug_pred_arch}, we compare the cosine similarity between the embedding of the augmented input $z'_m$ and the equivariant embedding predicted by the predictor $\hat{z}_m$ for each architecture.
The proposed attention-based architecture predicts displacement more precisely than others. 
Moreover, the well-trained equivariance leads to better performance in downstream tasks.

\begin{table}[t!]
\setlength{\tabcolsep}{8pt}
    \caption{Zero-shot retrieval results on AudioSet and audio-visual event classification performance on AudioSet-20K with the different training strategies.}
    \centering
    \label{abl:train_stratey}
    \begin{adjustbox}{max width=\linewidth}
    \begin{tabular}{lccccc}
        \toprule
        &\multicolumn{2}{c}{ZS Retrieval} & \multicolumn{3}{c}{Fine-Tuning (mAP)} \\ 
        \cmidrule(lr){2-3} \cmidrule(ll){4-6}
        Training Strategy     & V2A   & A2V   & A    & V     & A-V\\
        \midrule
        Two-Stage & 27.0 & 25.6 & 41.0 & 23.1 & 43.8 \\
        Alternating iterative  & 28.5 & 29.0 & 41.6 & 24.8 & 44.7 \\
        Joint & \textbf{30.1} & \textbf{29.6} &  \textbf{42.4} & \textbf{25.7} & \textbf{46.6} \\
        \bottomrule
    \end{tabular}
    \end{adjustbox}
    \vspace{-1em}
\end{table}

\paragraph{Training Strategies.}

We also conduct an ablation study on different training strategies for our framework.
First, the two-stage training strategy divides the intra-modal equivariant representation learning phase and the inter-modal representation learning phase.
In the first stage, the audio and visual encoders are trained using the intra-modal loss (Eq.~\ref{loss:intra}), and in the second stage, both encoders undergo combined training through inter-modal loss (Eq.~\ref{loss:inter}). 
Another training strategy employs an alternating iterative weight update of intra-modal loss and inter-modal loss.
For a fair comparison, all strategies are trained with the same epoch.
In Table~\ref{abl:train_stratey}, the two-stage and alternating iterative training strategies provide suboptimal results compared to the joint training. 
The result demonstrates joint training allows sufficient learning of intra-modal and inter-modal representations simultaneously without forgetting salient information.

\begin{table}[t!]
    \setlength{\tabcolsep}{10pt}
    \caption{Zero-shot retrieval results on AudioSet and audio-visual event classification performance on AudioSet-20K with different loss functions.}
    \label{sup:deno_pos}
    \centering
    \begin{adjustbox}{max width=\columnwidth}
    \begin{tabular}{lccccc}
    \toprule
    &\multicolumn{2}{c}{Zero-shot Retrieval} & \multicolumn{3}{c}{Fine-Tuning (mAP)} \\ 
    \cmidrule(lr){2-3} \cmidrule(ll){4-6}
    Intra-modal Loss     & V2A   & A2V   & A    & V     & A-V\\
    \midrule
    without pos (Eq.~\ref{loss:equimod_ell})  & 21.9 & 21.0 & 39.5 & 22.0 & 42.6 \\
    with pos (Eq.~\ref{loss:intra_ell})  &\textbf{30.1} & \textbf{29.6} &  \textbf{42.4} & \textbf{25.7} & \textbf{46.6}  \\
    \bottomrule
    \end{tabular}
    \end{adjustbox}
\end{table}
\paragraph{Equivariant Loss Functions.}
Table~\ref{sup:deno_pos} shows the results of using different equivariant loss functions for intra-modal pairs. 
The first row excludes the positive pair in the denominator according to Equation~\ref{loss:equimod_ell}, while the second row includes it as in Equation~\ref{loss:intra_ell}. 
When Equation~\ref{loss:intra_ell} is used for the intra-modal contrastive loss, the weight updates due to hard positives are relatively larger than those due to easy positives. 
In the context of equivariant contrastive learning, learning from hard positive with the transformation predictor is more effective, leading to better representation quality. 
Consequently, the semantically rich intra-modal representation promotes effective learning of audio-visual correspondence and audio-visual joint representations.
The experimental results support our hypothesis.
For further analytical insights, refer to Appendix~\ref{sup:loss}.

\subsection{Qualitative Results}
The evaluation results and ablation studies indicate that the proposed representative embedding effectively captures the salient information of audio-visual inputs.
We also explore how the relationship between the original embedding and the representative embedding is manifested in the input space. 
To this end, 1,000 augmentation vectors are randomly generated for each sample and then the equivariant embedding nearest to the centroid is selected as the substitute for the centroid. 
The corresponding transformation for this equivariant embedding is likely the one that best represents the image in the input space. 
The qualitative results in Figure~\ref{fig:qual} further validate our hypothesis, specifically by highlighting the parts of the image corresponding to the paired audio.
In addition, the zero-shot sound source localization task is performed on the original images and the images selected by the above method. 
While the attention map on the original images is dispersed to parts beyond the main elements, the attention for the augmented images, which correspond to the substitute embeddings of the centroids, is precisely and intensively concentrated on the areas that make the sound.

\section{Conclusion} 
In this paper, we propose EquiAV, a novel self-supervised audio-visual contrastive learning framework that incorporates the principle of equivariance.
EquiAV overcomes the limitations of applying augmentation in multi-modal representation learning by deriving a representative embedding with a proposed shared transformation predictor.
Extensive quantitative and qualitative results support the validity of our approach.
Furthermore, our approach can be adopted in other multi-modal domains, such as vision-language.
In particular, a multi-modal dataset with paired data augmentations enables the diverse approaches of applying equivariance in multi-modal representation learning.
We leave it as future work and we hope that EquiAV will help to push the boundaries in multi-modal representation learning.


\section*{Impact Statement}
This paper presents work whose goal is to advance the field of Machine Learning. There are many potential societal consequences of our work, none of which we feel must be specifically highlighted here.
\section*{Acknowledgements}

This work was supported in part by Institute of Information \& communications Technology Planning \& Evaluation (IITP) grant funded by the Korea government (MSIT, No. 2022-0-00184, Development and Study of AI Technologies to Inexpensively Conform to Evolving Policy on Ethics), by the National Research Foundation of Korea grant funded by the Korean government (MSIT, RS-2023-00212845), and by the Information Technology Research Center (ITRC) support program (IITP-2024-RS-2023-00259991) supervised by Institute of Information \& communications Technology Planning \& Evaluation (IITP).

\bibliography{icml2024/main}
\bibliographystyle{icml2024}

\newpage
\appendix
\onecolumn
\section{Appendix}
\appendix

\setcounter{table}{0}
\renewcommand{\thetable}{\Alph{table}}

\setcounter{figure}{0}
\renewcommand{\thefigure}{\Alph{figure}}

\setcounter{equation}{0}
\renewcommand{\theequation}{\Alph{equation}}

\setcounter{algorithm}{0}
\renewcommand{\thealgorithm}{\Alph{algorithm}}

\section{Additional Ablation Studies}
\label{sup:abl}
\paragraph{Weight-Scales of Loss Function.}
To understand how audio-visual contrastive learning is influenced by learning from each data modality, we explored the effects of setting $\lambda_{a,intra}$ and $\lambda_{v,intra}$ in Table~\ref{sup:lambda}. When self-supervision is derived solely from the audio domain (i.e. $\lambda_{v}=0$), it yielded the highest performance in audio-only tasks but showed suboptimal performance in visual-only and audio-visual tasks, and vice versa. Furthermore, adjusting the values of $\lambda_{a}$ and $\lambda_{v}$ to be either higher or lower relative to $\lambda_{inter}$ also resulted in relatively lower performance. Consequently, we found that using equal $\lambda$ values for both modalities and inter-modality learning produced balanced results.
\begin{table}[h]
    \setlength{\tabcolsep}{10pt}
    \caption{Audio-visual event classification performance on AudioSet-20K with varying weights scales of loss function.}
    \label{sup:lambda}
    \centering
    \begin{adjustbox}{max width=\linewidth}
    \begin{tabular}{ccccccc}
        \toprule
        & & &  \multicolumn{3}{c}{Fine-Tuning} \\ 
        \cmidrule(ll){4-6}
        $\lambda_{inter}$ & $\lambda_{a;intra}$ & $\lambda_{v;intra}$ &  A & V & A-V\\
        \midrule
        1 & 1 & 0 &  \textbf{42.9} & 24.0 & 43.4 \\
        1 & 0 & 1 &  41.2 & \textbf{26.3} & 43.2 \\
        1 & 2 & 2 &  42.2 & 25.5 & 45.5 \\
        1 & 0.5 & 0.5 & 41.3 & 25.1 & 46.0\\
        1 & 1 & 1 & 42.4 & 25.7 & \textbf{46.6} \\
        \bottomrule
    \end{tabular}
    \end{adjustbox}
\end{table}

\paragraph{Computational Cost \& Scalability.}
Table~\ref{sup:cost} reports the computational costs required by generating equivariant representations and Table~\ref{sup:scale} shows how the size of the pre-training dataset affects performance on downstream tasks. The results demonstrate the computational efficiency and scalability of our framework. 
\begin{table}[!h]
\setlength{\tabcolsep}{10pt}
    \caption{Computation cost for generating equivariant representations in audio and visual domains.}
    \vspace{-0.3em}
    \centering
    \label{sup:cost}
    \begin{adjustbox}{max width=\linewidth}
    \begin{tabular}{lll}
        \toprule
        $\#$ representations  & Audio(GFLOPs) & Visual(GFLOPs) \\ 
        \midrule
        Baseline	& 97.80	& 35.20 \\
        1		& 98.21 (+0.41)		& 35.36 (+0.16) \\
        4		& 98.23 (+0.43)		& 35.38 (+0.18) \\
        8		& 98.25 (+0.45)		& 35.40 (+0.20) \\
        16		& 98.30 (+0.50)		& 35.44 (+0.24) \\
        \bottomrule
    \end{tabular}
    \end{adjustbox}
    \vspace{-0.7em}
\end{table}

\begin{table}[!h]
    \setlength{\tabcolsep}{10pt}
    \caption{Zero-shot retrieval results on AudioSet and audio-visual event classification performance on AudioSet-20K with the variants of pre-training datasets.}
    \label{sup:scale}
    \centering
    \begin{adjustbox}{max width=\columnwidth}
    \begin{tabular}{lccccc}
    \toprule
    &\multicolumn{2}{c}{Zero-shot Retrieval} & \multicolumn{3}{c}{Fine-Tuning (mAP)} \\ 
    \cmidrule(lr){2-3} \cmidrule(ll){4-6}
    Pre-training Dataset     & V2A   & A2V   & A    & V     & A-V\\
    \midrule
    AudioSet-20K  & 3.9 & 4.2 & 33.7 & 17.3 & 35.2 \\
    VGGSound  & 11.2 & 10.5 & 37.6 & 21.3 & 40.8  \\
    AudioSet-2M  & \textbf{30.1} & \textbf{29.6} &  \textbf{42.4} & \textbf{25.7} & \textbf{46.6} \\
    \bottomrule
    \end{tabular}
    \end{adjustbox}
\end{table}

\section{Analysis on Equivariant Loss Functions}
\label{sup:loss}
As mentioned in the ablation study on Table~\ref{sup:deno_pos}, our equivariant loss function differs from the equivariant contrastive loss function used in EquiMod~\cite{devillers2023equimod}, regarding whether a positive pair similarity is included in the denominator. 
The equivariant loss of EquiMod is the same as applying an indicator function to both summation terms in the denominators of Equation~\ref{loss:intra_ell}, which can be represented as follows:
\begin{equation}\label{loss:equimod_ell}
    \ell_{\text{EquiMod}}(\hat{z}, z') = -\frac{1}{N} \sum_{i=1}^{N} \log \frac{\exp (\text{sim}(\hat{z}_{i}, {z'}_{i})/\tau)}{\sum_{j=1}^{N} \mathbbm{1}_{[j \neq i]} \left[ \exp(\text{sim}(\hat{z}_{i}, {z'}_{j})/\tau) + \exp(\text{sim}(\hat{z}_{i}, \hat{z}_{j})/\tau)\right]},
\end{equation}

\begin{equation}\label{loss:equimod}
   \mathcal{L}_{\text{EquiMod}} = \frac{1}{2} (\ell_{\text{EquiMod}}(\hat{z}, z') + \ell_{\text{EquiMod}}(z', \hat{z})).
\end{equation}

Consider intra-modal training batch embeddings as $\{{z}_i\} = \{\hat{z}_i\} \cup \{{z'}_i\}$.
For each $i$-th embedding, let's denote the positive sample as $p_i$ and the set of negative samples as $N_i$. Then, the equivariant loss terms of EquiMod and EquiAV can be simply rewritten as

\begin{equation}
    \mathcal{L}_{i}^{\text{EquiMod}} = -\log \frac{s_{i,p_i}}{\sum_{n \in N_i}s_{i,n}},
\end{equation}

\begin{equation}
    \mathcal{L}_{i}^{\text{EquiAV}} = -\log \frac{s_{i,p_i}}{s_{i,p_i} + \sum_{n \in N_i}s_{i,n}},
\end{equation}

where $s_{i,j}=\exp(\text{sim}(z_i, z_j)/\tau)$. 
Differentiating the above equations with respect to $s_{i,p_i}$ yields the following expressions:

\begin{equation}\label{equ:der_equimod}
\frac{\partial \mathcal{L}_{i}^{\text{EquiMod}}}{\partial s_{i,p_i}} = -\frac{1}{s_{i, p_i}} ,
\end{equation}
\vspace{-0.5em}
\begin{align} \label{equ:der_equiav}
\begin{split}
\frac{\partial \mathcal{L}_{i}^\text{EquiAV}}{\partial s_{i,p_i}} &= -\frac{\sum_{n \in N_i} s_{i, n}}{ s_{i, p_i}  \left(s_{i, p_i}+\sum_{n \in N_i} s_{i, n}\right)} \\ 
&= \frac{\partial \mathcal{L}_{i}^\text{EquiMod}}{\partial s_{i,p_i}} \cdot \frac{\sum_{n \in N_i} s_{i, n}}{\left(s_{i, p_i}+\sum_{n \in N_i} s_{i, n}\right)}
\end{split}
\end{align}

When we compare the EquiAV loss to the EquiMod loss through Equ.\ref{equ:der_equiav} and Equ.~\ref{equ:der_equimod}, the EquiAV loss puts relatively more weight on hard positive compared to easy positive.
This becomes particularly advantageous when stronger augmentations lead to an increased frequency of hard positives. 
Training with more hard positives and applying equivariance substantially enhances the model's capability to comprehend detailed features in single-modal contexts. 
As a result, models enhanced with this function show improved performance and effectively integrate different modalities.

\section{Implementation Details}
\label{sup:implementation}
\paragraph{Datasets} 
We utilize two prominent audio-visual datasets for our experiments: AudioSet~\cite{gemmeke2017audio} and VGGSound~\cite{chen2020vggsound}. 
AudioSet comprises 2 million 10-second YouTube clips, designed to classify events into 527 distinct classes, with each data having multiple labels. 
We download 1,893,278 clips for the AudioSet-2M, 21,074 clips for the AudioSet-20K, and 19,446 clips for evaluation. Particularly, AudioSet-20K is a subset of AudioSet-2M. 
VGGSound includes 200,000 10-second YouTube clips, encompassing 309 classes. 
The training and test splits of VGGSound consist of 183,730 and 15,446 downloaded clips, respectively.
Unlike AudioSet, it has only one label for each clip.
For the zero-shot retrieval evaluation, we collect 1,722 and 1,545 clips from AudioSet and VGGSound's evaluation set respectively. In addition, MSR-VTT~\cite{xu2016msr} test set is also used in the zero-shot retrieval task.

\paragraph{Input Pre-processing.}
We follow AST~\cite{gong21binterspeech} and ViT~\cite{dosovitskiy2021an} for pre-processing of audio and visual inputs, respectively.
For audio, each 10-second audio waveform is transformed into a sequence of 128-dimensional log Mel filterbank features by using a 25-ms Hanning window and a 10-ms hop size, resulting in a 1024(time) $\times$ 128(frequency) spectrogram.
For visual inputs, 10 frames are uniformly extracted from each 10-second video, and one frame is randomly selected as the input.
Then the audio spectrogram and the video frame are tokenized to 16 $\times$ 16 patches and fed to the audio and visual encoders.
In terms of input data augmentation for the visual modality, the typical visual-based augmentations are used; Random Resized Crop (RRC), Color Jitter (CJ), Gaussian Blur (GB), Horizontal Flip (HF), and Gray Scale (GS).
On the other hand, both audio- and visual-based augmentations are applied to the audio spectrogram. 
Specifically, SpecAugment (SA)~\cite{park2019specaug} and Time Shifting (TS) are utilized as audio-based augmentations.
Meanwhile, the same augmentations pool used for the visual modality excluding GS is applied to audio spectrograms, as the spectrogram has only one channel.

\paragraph{Augmentation Parameterization.}
\label{sup:aug_param}
Audio and visual augmentations as well as their parameterization used in this work are listed as follows:

\begin{itemize}
\setlength\itemsep{0.5em}
\item \textit{Random resized crop} (4 elements for both audio \& visual): \textit{x} and \textit{y} coordinates of the top-left point, as well as the width and height of the crop ($<$0, 0, 0, 0$>$, is used as a default encoding).
\item \textit{Color jitter} (8 elements for visual \& 4 elements for audio): the jitter factors for brightness, contrast, saturation, and hue of video frames, as well as the order of the application of transformation. We use the following mapping to encode the order of transformation: $\{$0: brightness, 1: contrast, 2: saturation, 3: hue$\}$. For instance, an encoding $<$2, 1, 3, 0$>$ indicates that the saturation jitter is first applied, and then contrast, hue, and brightness ($<$0, 1, 2, 3$>$ is used as default). On the other hand, we only use brightness and contrast jitters for audio spectrograms, as the audio spectrograms are originally grayscale. Then, we use the following mapping to encode the order of jitter transformation: {0: brightness, 1: contrast} ($<$0, 1$>$ is used as default).
\item \textit{Gaussian blur} (1 element for both audio \& visual): the value of \textsigma{} for Gaussian blurring kernel (0 as default).
\item \textit{Horizontal flip} (1 element for both audio \& visual): 1 if an image or an audio spectrogram is horizontally flipped and 0 otherwise.
\item \textit{Grayscale} (1 element, for visual only): 1 if an image is converted to grayscale and 0 otherwise.
\item \textit{Random time shifting} (1 element, for audio only): the value of temporal shift of the audio spectrogram.
\item \textit{SpecAug}~\cite{park2019specaug} (4 elements, for audio only): starting and end points of the masking along the time and frequency axis of the audio spectrogram.
\end{itemize}

All augmentations except random resized crop (which is always applied) are applied with pre-defined probability.
Therefore for each augmentation, we add an element, whose value is 1 when the augmentation is actually applied and 0 otherwise, to a parameterized vector.
Consequently, the audio and visual augmentations are encoded into 24-dimensional and 18-dimensional vectors respectively.

\newpage
\section{Hyperparameter Details}
\label{sup:hyper}
The hyperparameter settings used in this paper are listed in Table~\ref{sup:config}.

\begin{table}[!thb]
\caption{Hyperparameters used in pre-training and fine-tuning phase.}
\centering
\label{sup:config}
\begin{adjustbox}{max width=1.0\columnwidth,center}
{\begin{tabular}{lcccccccccc}
\toprule
Stage                  & \multicolumn{1}{c}{Pre-training} & \multicolumn{9}{c}{Fine-Tuning} \\
\cmidrule(lr){2-2} \cmidrule(lr){3-11}

Dataset                & AudioSet-2M      & AudioSet-20K   & AudioSet-2M   & VGGSound & ESC-50 & SPC-v1  & FSD-50K & UCF101 & HMDB51 & K400\\
\midrule
Optimizer              & \multicolumn{10}{c}{AdamW} \\
Optimizer momentum     & \multicolumn{10}{c}{$\beta_1$=0.9,\qquad $\beta_2$=0.95} \\
Weight decay           & \multicolumn{10}{c}{1e-5} \\
Learning rate scheduler& \multicolumn{10}{c}{half-cycle cosine annealing \citep{loshchilov2016sgdr}} \\
Initial learning rate  & \multicolumn{10}{c}{1e-6} \\
Peak learning rate     & 1e-4   & 1e-4   & 1e-4   & 1e-4 & 5e-4 & 1e-3 & 5e-4 & 1e-3 & 5e-4 & 5e-4 \\
Warm-up epochs         & 2        & 1         & 1       & 1 & 4 & 1  & 4 & 0 & 0 & 4 \\
Epochs                 & 20         & 50       & 50      & 50  & 60 & 10  & 60 & 100 & 100 & 100  \\
Batch size             & 256         & 256       & 512      & 256  &  64  & 256  & 64 & 32 & 32 & 512  \\
Class Balancing Weight & No         & No       & Yes     & Yes    & No & No & No & No & No & No\\
Mixup                  & No         & Yes      & Yes     & Yes   & No & Yes & No & No & No & Yes  \\
Loss Function          & EquiAV Loss (Eq.~\ref{loss:equiav})        & BCE      & BCE     & CE      & CE & BCE & BCE & CE & CE & CE  \\
Temperature ($\tau$)     & 0.07      & -         & -        & -   & - & - & - & - & - & -   \\
Input Norm Mean        & -4.346     & -4.346   & -4.346  & -4.956   & -6.627 & -6.702 & -6.627 & - & - & - \\
Input Norm STD         & 4.332      & 4.332    & 4.332   & 4.486  & 5.359 & 5.448 & 5.359 & - & - & -   \\ 
GPUs                   & 8 A6000    & 8 A5000   & 8 A6000 & 8 A5000 & 8 A5000 & 8 A5000 & 8 A5000 & 8 A5000 & 8 A5000 & 8 A5000 \\

\bottomrule
\end{tabular}}
\end{adjustbox}
\end{table}

\newpage

\end{document}